\documentclass[11pt]{article}
\usepackage{amsmath}
\usepackage{algorithm}
\usepackage{algorithmic}
\usepackage{amsfonts}
\usepackage{tabularx}

\usepackage[english]{babel}

\usepackage[letterpaper,top=2cm,bottom=2cm,left=3cm,right=3cm,marginparwidth=1.75cm]{geometry}

\usepackage[letterpaper,top=2cm,bottom=2cm,left=3cm,right=3cm,marginparwidth=1.75cm]{geometry}

\usepackage{amsmath}
\usepackage{graphicx}
\usepackage[colorlinks=true, allcolors=blue]{hyperref}

\usepackage{authblk}

\title{Towards \textit{more} Contextual Agents: An extractor-Generator Optimization Framework}
\author{
Mourad Aouini$^{1,2}$, Jinan Loubani$^{1}$\\
{\small $^1$CGI,}
{\small $^2$CNRS}
}

\date{}

\begin{document}
\maketitle

\begin{abstract}
Large Language Model (LLM)-based agents have demonstrated remarkable success in solving complex tasks across a wide range of general-purpose applications. However, their performance often degrades in context-specific scenarios, such as specialized industries or research domains, where the absence of domain-relevant knowledge leads to imprecise or suboptimal outcomes. To address this challenge, our work introduces a systematic approach to enhance the contextual adaptability of LLM-based agents by optimizing their underlying prompts—critical components that govern agent behavior, roles, and interactions. Manually crafting optimized prompts for context-specific tasks is labor-intensive, error-prone, and lacks scalability. In this work, we introduce an Extractor-Generator framework designed to automate the optimization of contextual LLM-based agents. Our method operates through two key stages: (i) feature extraction from a dataset of gold-standard input-output examples, and (ii) prompt generation via a high-level optimization strategy that iteratively identifies underperforming cases and applies self-improvement techniques. This framework substantially improves prompt adaptability by enabling more precise generalization across diverse inputs, particularly in context-specific tasks where maintaining semantic consistency and minimizing error propagation are critical for reliable performance. Although developed with single-stage workflows in mind, the approach naturally extends to multi-stage workflows, offering broad applicability across various agent-based systems. Empirical evaluations demonstrate
that our framework significantly enhances the performance of prompt-optimized agents, providing a structured and efficient approach to contextual LLM-based agents.
\end{abstract}

\section{Introduction}

Large Language Model (LLM)-based agents have achieved remarkable success in tackling complex tasks across diverse, general-purpose applications, demonstrating versatility in understanding, reasoning, and generating coherent responses across a wide array of domains \cite{wang2023survey}\cite{zhang2024comprehensive}\cite{lala2023paperqa}. However, when applied to context-specific scenarios, such as specialized industries or research domains, LLM performance often suffers from a lack of domain-specific knowledge, resulting in less accurate and less reliable outcomes for tasks requiring detailed and precise understanding.  For instance, \cite{lehman2023clinical} found that smaller, specialized clinical models can substantially outperform larger LLMs in tasks requiring specific domain knowledge. To tackle this challenge, we present a systematic method for improving the contextual adaptability of LLM-based agents. Our approach focuses on optimizing the agents defined by prompts that control their behavior, roles, and interactions, ensuring more accurate performance in domain-specific tasks. Another key advantage of our proposed approach is its ability to mitigate the problem of prompt sensitivity highlighted by \cite{verma2024brittle}, resulting in agents that maintain stable performance across varying prompts and task conditions.

Classical prompting techniques, such as chain-of-thought prompting \cite{wei2022chain}, program-of-thoughts prompting \cite{chen2022program}, and multistage pipeline architectures \cite{wu2022ai} \cite{khattab2022demonstrate} \cite{schlag2023large}, have significantly improved the reasoning capabilities of LLMs. However, these methods often remain insufficient for achieving optimal performance in LLM-based agents, particularly in complex, context-dependent tasks. Current approaches to constructing LLM-based agents predominantly rely on manual prompt engineering, involving predefined templates and human-crafted modifications—a process that is both time-intensive and prone to suboptimal outcomes \cite{austin2024gradsum}. Recent advancements in declarative language model pipelines \cite{khattab2024dspy}\cite{opsahl2024optimizing}\cite{wan2024teach}\cite{wan2025manyshot} aim to optimize critical prompt components, such as instructions and demonstrations, yet these approaches do not cover other critical components that define prompt-optimized agents behavior. This gap underscores the necessity of developing a systematic and automated framework for optimizing LLM-based agents, which is crucial to enhance their efficiency, adaptability, and reliability across diverse, context-specific applications.

In this article, we address the problem of optimizing LLM-based agents by formulating it as a structured learning task, and by decomposing the prompt-optimized agent into several components. We propose an Extractor-Generator framework that automates optimizing contextual LLM-based agents by systematically refining their performance. In the first stage, a feature extraction module analyzes a dataset of gold-standard input-output pairs to identify key contextual features. These extractions are performed in parallel on a small dataset. In the second stage, a prompt-generation mechanism employs an iterative optimization strategy, detecting underperforming cases and applying self-improvement techniques. Unlike prior research, which focuses on manually refining prompts or leveraging external knowledge sources, our framework automatically refines agents components based on iterative feedback. This two-step approach enhances the effectiveness of prompt-optimized agents, allowing the model to generalize better across diverse inputs while minimizing errors in context-specific applications. By iteratively refining the performance of feedback-based prompts, the framework reduces error propagation, thereby improving the reliability of LLM-based agents. While initially designed for single-stage workflows, its modular structure inherently supports extension to multistage processes, demonstrating potential applicability across a wide range of agent-based systems. Empirical results show that our framework substantially improves the performance of prompt-optimized agents, offering a systematic and efficient method to enhance the capabilities of contextual LLM-based agents.
    
\section{Designing Extractor-Generator Framework}
\begin{figure}[!ht]
\centering
\includegraphics[width=\linewidth]{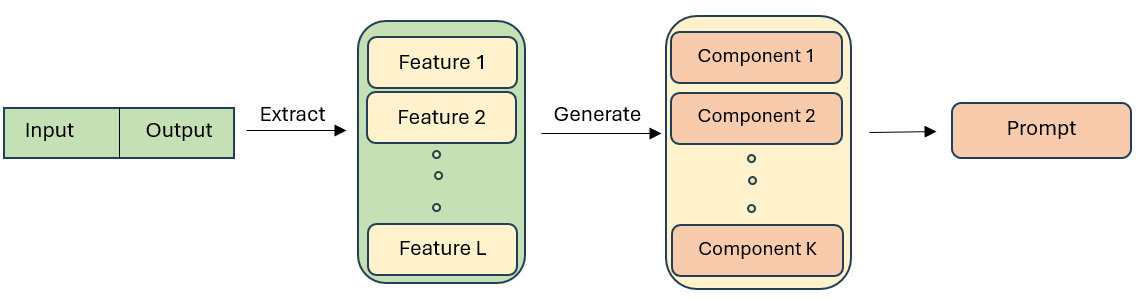}
\caption{\label{ext_gen}Architecture of Features Extraction and Prompt-Components Generation.}
\end{figure}

\noindent In this section, we present the Extractor-Generator Framework, providing a comprehensive formulation of the underlying optimization problem. The framework is designed to automate and enhance the performance of LLM-based agents through a systematic, data-driven process. Our approach is divided into two key stages:

\begin{enumerate}
    \item \textbf{Features Extraction}: This stage involves analyzing task-specific data to generate a matrix $\mathcal{M}$ of feature vectors. These features capture essential patterns and contextual signals that influence agent behavior and performance.
    
    \item \textbf{Prompt-Components Generation}: Using the extracted feature representations, this stage focuses on constructing and refining prompt components. The goal is to optimize the agent’s behavior by generating contextually relevant instructions, demonstrations, and other prompt elements that guide the model's decision-making.
\end{enumerate}

Formally, given the matrix of feature vectors $\mathcal{M}$ and an LLM-based agent \(\Psi\), the optimization problem can be mathematically defined as follows:

\[\Psi_\mathbf{v}^*=
\arg\max_{\mathbf{v} \in \mathcal{M}} \frac{1}{\mid \mathcal{IO}\mid}\sum_{(x,y)\in \mathcal{IO}}{}f(\Psi_\mathbf{v}(x),y) 
\]

\noindent where $f$ is an objective function and $\mathcal{IO}$ is the set of gold input-output pairs. Figure \ref{ext_gen} shows the overall architecture of features extraction and prompt generation.

\subsection{Features Extraction}
\begin{figure}[!ht]
\centering
\includegraphics[width=\linewidth]{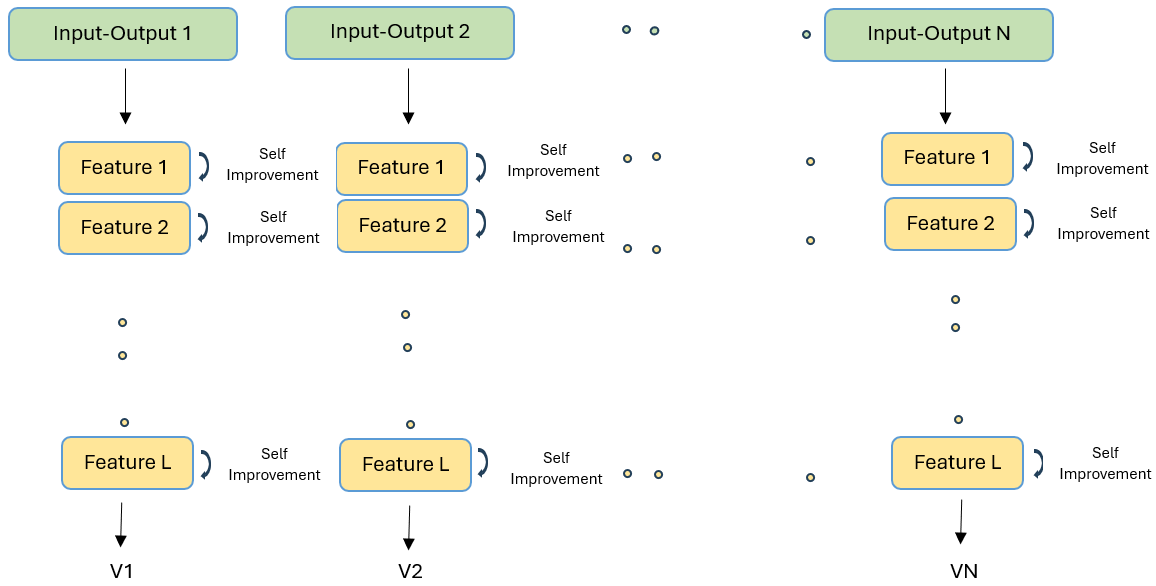}
\caption{\label{extraction}Features Extraction on gold data.}
\end{figure}


\noindent Given a small set of gold-standard input-output pairs $\mathcal{IO}$, we extract the key contextual features using an Extractor-LLM. This extractor operates as a multi-agent system with a fully distributed topology, where each agent is specialized in capturing a distinct feature dimension. While all agents utilize the same underlying Extractor-LLM, multiple extractors can be employed to enhance robustness and coverage. For each extracted feature, a self-improvement mechanism is applied, as illustrated in Figure \ref{extraction}, to iteratively refine the feature representation. Consequently, each input-output pair is transformed into a feature vector that encodes task-relevant contextual information. The resulting contextual feature vectors serve as a foundation for generating the core prompt components, as described in the following section.

\begin{algorithm}[!ht]
\caption{Optimize Prompt $P$ of LLM-based Agent $\Psi$}
\begin{algorithmic}[1]
\STATE \textbf{Input:}
\STATE \hspace{1cm} $\Psi$: LLM-based Agent
\STATE \hspace{1cm} $\mathcal{IO}$: Dataset of gold examples (input-output pairs)
\STATE \hspace{1cm} $P$: Prompt viewed as a function of its $L$ components
\STATE \hspace{1cm} $E$: Extractor LLM
\STATE \hspace{1cm} $G$: Generator LLM
\STATE \hspace{1cm} $B$: Mini batch size
\STATE \hspace{1cm} $T$: Number of batches
\vspace{0.5em}
\STATE \textbf{Output:}
\STATE \hspace{1cm} Optimized prompt $P_{\text{opt}}$
\vspace{0.5em}
\STATE \textbf{Extract features:}
\hspace{1cm} \FOR{$(x,y)\in \mathcal{IO}$}
    \STATE Use $E$ to extract the features vector $V_{x,y}$ of $(x,y)$
    \STATE Self-improve $V_{x,y}$ of $(x,y)$ using $E$ 
    \ENDFOR
\STATE Consider the matrix $\mathcal{M}_{N, L}$ whose rows are the vectors $V_{x,y}$
\vspace{0.5em}
\STATE \textbf{Generate prompt:}
\FOR{$t = 1$ to $T$}
\STATE Consider a sub-matrix $\mathcal{M}_{B, L}$ of $B$ rows of $\mathcal{M}_{N, L}$
\STATE Call $G$ to initialize $P_t$ using the columns of $\mathcal{M}_{B, L}$
\STATE Self-improve $P_t$ using $G$
\STATE Evaluate the performance $s_t$ of $P_t$ on $\mathcal{IO}$
\ENDFOR
\STATE Consider the prompt $P_b$ with the best performance $s_b$
\STATE Update $P_b$ with its corresponding $s_b$ using Algorithm \ref{alg2}
\STATE \textbf{Return:} Optimized prompt $P_{\text{opt}}$
\end{algorithmic}\label{alg1}
\end{algorithm}

\subsection{Components Generation}

\begin{figure}[!ht]
\centering
\includegraphics[width=\linewidth]{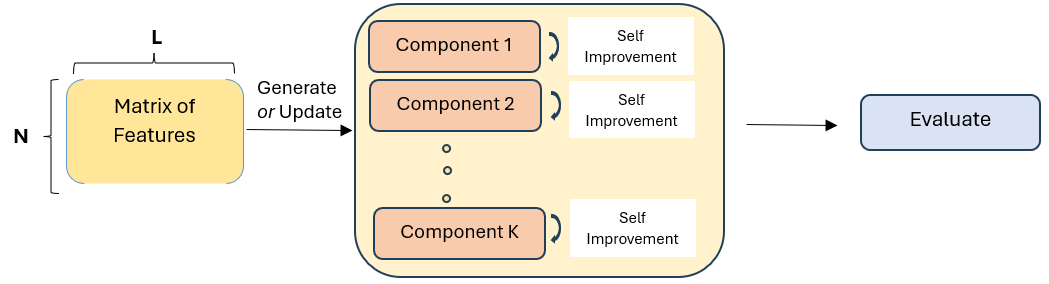}
\caption{\label{generation}Components Generation using Extracted Features and Self-Improvement.}
\end{figure}

We consider the matrix $\mathcal{M}_{N, L}$ whose $N$ rows are defined by the vectors $\{V_{x,y}\}_{(x,y)\in \mathcal{IO}}$ of extracted features, where $N$ is the dimension of the data and $L$ is the features dimension. A prompt $P_t$ is then generated iteratively using sub-matrices $\mathcal{M}_{B, L}$, randomly sampled batches of rows from $\mathcal{M}_{N, L}$, as shown in Algorithm \ref{alg1}. Each prompt $P_t$ is initialized and self-improved by a Generator LLM-based agent $G$, and its performance $s_t$ is evaluated on the dataset. The best-performing prompt $P_b$, based on its evaluation $s_b$, is further optimized using an additional refinement step shown in Algorithm~\ref{alg2}. At each iteration, it evaluates the prompt’s effectiveness on individual input-output pairs $(x,y)$ using an evaluation metric $\mu$. If $\mu$ indicates underperformance, i.e. if $\mu(P_i, \Psi(x),y)< \lambda$ for some threshold $\lambda$, the prompt is updated and self-improved using $G$. This process repeats up to a few number of times (three times for example), or until the threshold is met, to ensure self-improvement . The overall performance $s_i$ of the updated prompt is then calculated for the entire dataset. If the new performance $s_i$ surpasses the current best score $s$, the prompt and score are updated accordingly. After a maximum number of iterations, the optimal prompt $P_{\text{opt}}$ with the highest performance is returned. Figure \ref{generation} shows the steps followed in every iteration.

\begin{algorithm}[!ht]
\caption{Update Prompt $P$ of LLM-based Agent $\Psi$}
\begin{algorithmic}[1]
\STATE \textbf{Input:}
\STATE \hspace{1cm} $\Psi$: LLM-based Agent
\STATE \hspace{1cm} $\mathcal{IO}$: Dataset of gold examples (input-output pairs)
\STATE \hspace{1cm} $P$: Initial prompt 
\STATE \hspace{1cm} $s$: Initial performance 
\STATE \hspace{1cm} $E$: Extractor LLM
\STATE \hspace{1cm} $G$: Generator LLM
\STATE \hspace{1cm} $\mu$: Evaluation metric
\STATE \hspace{1cm} $I$: Max iterations 
\vspace{0.5em}
\STATE \textbf{Output:}
\STATE \hspace{1cm} Optimized prompt $P_{\text{opt}}$
\vspace{0.5em}
\FOR{$i = 1$ to $I$}
    \FOR{$(x,y)\in \mathcal{IO}$}
        \FOR{$k=1$ to $3$}
            \IF{$\mu(P_i, \Psi(x),y)<\lambda$}
             \STATE Update $P_i$ using $G$ to represent $(x,y)$
            \STATE Self-improve $P_i$ using $G$
            \STATE compute and update $\mu(P_i, \Psi(x),y)$
            \ENDIF
        \ENDFOR
        \STATE Compute performance $s_i$ for $P_i$ on $\mathcal{IO}$
        \IF{$s_i>s$}
            \STATE $P\xleftarrow{} P_i$
            \STATE $s\xleftarrow{} s_i$
            \ENDIF
        \ENDFOR
\ENDFOR
\STATE \textbf{Return:} Optimized prompt $P_{\text{opt}}$
\end{algorithmic}\label{alg2}               
\end{algorithm}

\subsection{Self-Improvement}
We use a mechanism of self-improvement three times in our approach; features extraction in Algorithm~\ref{alg1}, first prompt generation in Algorithm \ref{alg1}, and updating and optimizing prompt in Algorithm \ref{alg2}. Self-improvement in a workflow refers to the process where the system iteratively enhances its own performance by leveraging previous results as feedback. This involves feeding the program’s past outputs back into itself, allowing it to analyze errors, refine its parameters, and generate improved iterations over time. In the context of language models, self-improvement can be achieved by using an auxiliary model or internal mechanisms to adjust prompts, fine-tune responses, or optimize decision-making based on past evaluations. By continuously learning from its own performance, the workflow gradually converges toward more accurate, coherent, and efficient outputs.



\subsection{Multistage Workflow}
The generalization of the proposed Extractor-Generator Framework from single-stage to multi-stage workflows involves extending the optimization process from individual agents to the entire multi-agent system. Initially, each agent undergoes optimization using the same iterative feature extraction and prompt-generation method to enhance task-specific performance. Once agent-level optimization is complete, the framework applies the same optimization strategy at the system level, ensuring that inter-agent interactions and collaborative performance are jointly optimized. This system-level optimization is guided by a predefined workflow topology, which dictates the structural arrangement and operational orchestration of agents within the multi-stage pipeline. Workflow topologies can follow various configurations depending on task requirements. Effective connectivity between agents is essential to facilitate the seamless exchange of contextual information, ensuring consistency and coherence across stages. By integrating agent-level and system-level optimization within a structured yet adaptable workflow, the framework enhances the robustness and efficiency of LLM-based agents in complex, multi-stage applications.

\section{Experiments}
To rigorously evaluate the effectiveness of our optimization framework for LLM-based agents, we conduct a series of experiments across five distinct application domains: finance, healthcare, e-commerce and retail, law, and cybersecurity. Each domain-specific dataset comprises 150 examples for optimization and 300 examples for testing, presented in the form of Frequently Asked Questions (FAQs) to ensure a diverse and contextually rich evaluation set.

We evaluate and compare our contextual agent with multiple prompting pipelines, including chain-of-thought (CoT) prompting, sequential CoT and self-consistency CoT. We consider the optimal versions of these pipelines with only instructions optimized by our Extractor-Generator Framework. The evaluation relies on an answer relevancy metric, assessed by a dedicated LLM-based judge agent. This agent evaluates the generated responses based on correctness, coherence, and informativeness, providing a consistent and systematic framework for performance assessment.

\begin{table}[ht]
\centering
\small
\renewcommand{\arraystretch}{1.2}
\begin{tabularx}{\linewidth}{|l|X|X|X|X|X|}
\hline
\textbf{Workflow/Domain} & \textbf{Finance} & \textbf{Health- care} & 
\textbf{E-commerce \& Retail} & \textbf{Law} & \textbf{Cyber- security} \\
\hline
\textbf{CoT} & 74.5\% & 72.3\% & 76.2\% & 67.1\% & 62.4\% \\
\hline
\textbf{Sequencial CoT} & 75.1\% & 70.5\% & 73.3\% & 69.4\% & 62.7\% \\
\hline
\textbf{Self-Consistency CoT} & 78.3\% & 72.6\% & 79.5\% & 69.2\% & 65.5\% \\
\hline
\textbf{Our Contextual Agent} & \textbf{87.4\%} & \textbf{82.7\%} & \textbf{88.1\%} & \textbf{81.3\%} & \textbf{79.6\%} \\
\hline
\end{tabularx}
\caption{Performance comparison of different workflows across various application domains.}
\label{tab:performance_comparison}
\end{table}

Table \ref{tab:performance_comparison} presents the performance evaluation of the different optimal prompting pipelines across the five domains mentioned above. The analysis aims to assess how effectively each technique enhances the model’s ability to generate contextually accurate, coherent, and domain-relevant responses in diverse, real-world tasks. The results indicate that our contextual agent consistently outperforms the baseline approaches—CoT, sequential CoT, and self-consistency CoT—across all domains. Notably, it achieves its highest performance in the E-commerce and retail domain with 88.1\% relevancy, highlighting its capacity to adapt to customer-centric contexts. The finance and healthcare domains also exhibit significant performance gains, with our contextual agent achieving 87.4\% and 82.7\%, respectively.

These findings demonstrate the effectiveness of the proposed framework in optimizing prompts by extracting task-relevant features and dynamically adapting components to enhance domain-specific performance. The consistent improvements observed across all domains further underscore the generalizability of the method and its robustness in diverse, context-specific scenarios.

\section{Conclusion}
In this work, we introduced the Extractor-Generator Framework, a systematic and scalable solution for optimizing LLM-based agents in context-specific tasks. By implementing a two-stage process—feature extraction and prompt generation—our approach overcomes the limitations of manual prompt engineering, which is often labor-intensive, error-prone, and suboptimal. The framework identifies task-relevant contextual features through a distributed multi-agent extraction process and applies an iterative self-improvement mechanism to generate optimized prompts.

Empirical evaluations demonstrate that the proposed method significantly enhances the performance of LLM-based agents by improving their ability to generalize across diverse contexts while maintaining semantic consistency. The framework’s adaptability extends naturally from single-stage to multi-stage workflows, highlighting its potential for broader applications in complex, dynamic environments.

The results of this study underscore the critical role of automated, context-aware prompt optimization in enhancing the effectiveness and reliability of LLM-based agents. By shifting from static, handcrafted prompts to a dynamic, data-driven optimization process, the Extractor-Generator Framework provides a robust foundation for more efficient and contextually intelligent language model applications across a wide range of domains.

\bibliographystyle{alpha}
\bibliography{main}

\end{document}